\documentclass[conference]{IEEEtran}
\usepackage{graphicx}
\usepackage{caption}
\usepackage{mathrsfs}
\usepackage{algorithm}  
\usepackage{algorithmicx}  
\usepackage{algpseudocode}
\usepackage{booktabs}
\usepackage{amsmath}
\usepackage{amssymb}
\usepackage{url}

\begin{document}

\title{Self-Imitation Learning for Robot Tasks with Sparse and Delayed Rewards}

\author{\authorblockN{Zhixin Chen}
\authorblockA{School of Mechanical Engineering and Automation, \\ Beihang University, Beijing 100191, China\\
Nanjing Research Institute of Electronics Technology,\\
Nanjing 210039, China\\
Email: zhixinc@buaa.edu.cn}
\and
\authorblockN{Mengxiang Lin}
\authorblockA{School of Mechanical Engineering and Automation, \\ Beihang University, Beijing 100191, China\\
Email: linmx@buaa.edu.cn\\
* Corresponding author
}
}

% avoiding spaces at the end of the author lines is not a problem with
% conference papers because we don't use \thanks or \IEEEmembership

%Homer Simpson\authorrefmark{2},
%Montgomery Scott\authorrefmark{3} and
%Eldon Tyrell\authorrefmark{4}}
%Telephone: (800) 555--1212, Fax: (888) 555--1212}
%\authorblockA{\authorrefmark{4}Tyrell Inc., 123 Replicant Street, Los Angeles, California 90210--4321}}

%\specialpapernotice{(Invited Paper)}

\maketitle

\begin{abstract}
The application of reinforcement learning (RL) in robotic control is still limited in the environments with sparse and delayed rewards.
In this paper, we propose a practical self-imitation learning method named Self-Imitation Learning with Constant Reward (SILCR).
Instead of requiring hand-defined immediate rewards from environments, our method assigns the immediate rewards at each timestep with constant values according to their final episodic rewards.
In this way, even if the dense rewards from environments are unavailable, every action taken by the agents would be guided properly.
We demonstrate the effectiveness of our method in some challenging continuous robotics control tasks in MuJoCo simulation and the results show that our method significantly outperforms the alternative methods in tasks with sparse and delayed rewards.
Even compared with alternatives with dense rewards available, our method achieves competitive performance.
The ablation experiments also show the stability and reproducibility of our method.
\end{abstract}


% for peerreview papers, inserts a page break and creates the second title.
% Will be ignored for other modes.
\IEEEpeerreviewmaketitle

\section{Introduction}
% no \PARstart
Reinforcement learning (RL) offers a promising way for robots to learn complex skills in a self-supervision manner \cite{Hwangbo2019LearningAA,realworldRL,Zhu2019DexterousMW}.
It is well known that the performance of RL depends crucially on the quality of reward functions.
However, the rewards in real-world robotic applications are extremely sparse and delayed \cite{alphazero,HER}.
Take a robotic navigation task as an example \cite{Choi2019DeepRL}, a robot with the limited field of view is tasked to reach a goal position.
To each move of the robot in the midway, no reward is offered until the final goal is reached.
Then a final reward is obtained indicating whether the navigation efforts succeed or not.
How to assign this final reward reasonably to all the moves in navigation is crucial for learning, since not only the final move but also the previous moves lead to the final bonus.
In such a sparse and delayed setting, standard RL methods would fail to learn due to the immeasurably huge amount of experiences required \cite{DiverseSIL}.

Many techniques have been proposed to cope with this issue \cite{curiosity,Li2020LearningHC,SIL,SIL2019}.
In this paper, we shall only be interested in self-imitation learning (SIL).
SIL is first introduced to improve the performance of advantage actor-critic (A2C) algorithm in hard exploration tasks \cite{SIL}.
More generally, the concept of SIL is used to address temporal credit assignment problem in reinforcement learning \cite{DiverseSIL,GASIL}.
Furthermore, policy optimization can also be formalized as a divergence minimization problem under the framework of self-imitation learning \cite{DiverseSIL}.

In practice, SIL concept is usually implemented as the auxiliary component of a RL learning process.
A variety of strategies have been developed to guide the agents with imitated behaviors in the standard RL iterations \cite{DiverseSIL,SIL}.
With the help of these good behaviors, the current policy of the agents as well as the historical good behaviors will be  continuously improved. 
In this way, self-imitation learning combines the environmental reward term and self-imitation term to take advantage from both.
However, timely feedback from environments is still indispensable for learning in existing SIL methods.

In this paper we propose a novel SIL method named Self-Imitation Learning with Constant Reward (SILCR).
In our self-imitation setting, the constant reward assignment \cite{SQIL} is applied to replace the assignment of the environmental temporal dynamics.
Specifically, all the actions in a trajectory are assigned with constant rewards instead of the immediate rewards required from environments, and the goodness of trajectories would be determined by their episodic rewards which are finally provided by the environment when an episode terminates.
In principle, such a constant reward assignment makes all the actions being guided in a proper way and 
would significantly relieve the credit assignment problem in robotic tasks with sparse and delayed rewards.
To the best of our knowledge, this is the first work that performs self-imitation learning with no need for immediate rewards.

We evaluate the effectiveness of our method in some challenging robotic continuous control tasks in MuJoCo simulation.
Following common practice in the literature, a sparse reward version is generated from the original control tasks with dense rewards to have a fair comparison.
The results show that our method significantly outperforms RL alternatives in sparse reward settings.
Even compared with RL and SIL alternatives with dense rewards, our method achieves a competitive performance.

The core contribution of this work is a practical self-imitation learning method for circumventing the need for immediate rewards defined manually.
The cost of defining a dense and meaningful reward function can be prohibitively expensive in real-world robotic applications.
In addition, our method achieves impressive reproducibility since there is only one additional hyperparameter introduced and the ablation experiments show that the performance of our method is insensitive to it.

\section{Proposed Method}
\subsection{Self-imitation learning framework}
\begin{figure}[thpb]
\centering
\includegraphics[width=0.95\columnwidth]{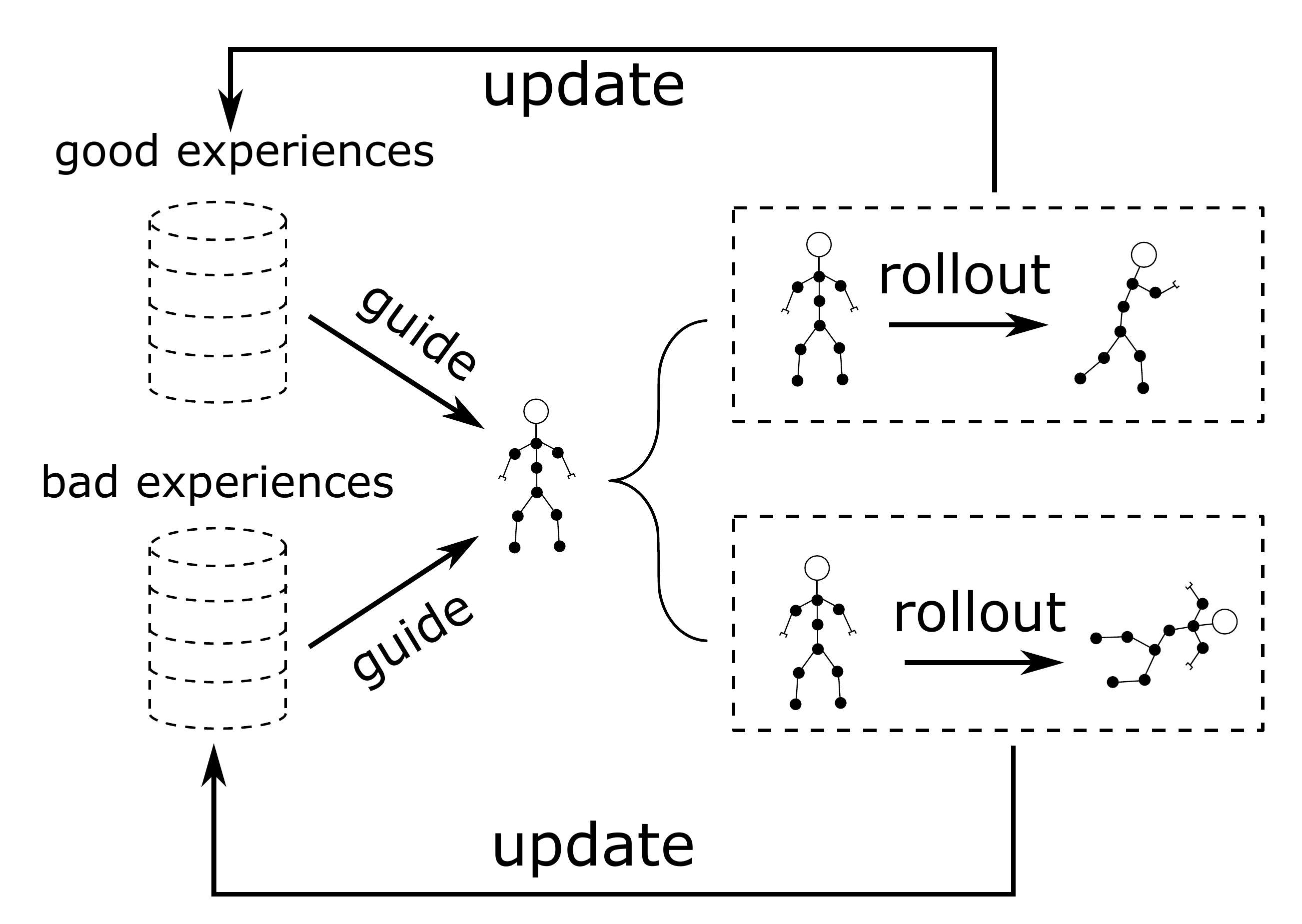} 
\caption{Our self-imitation learning framework for robot learning. The good (keep walking) and bad (fall down) experiences generated by the robot itself will further guide its subsequent behaviors.}
\label{intuition}
\end{figure}

In a sparse and delayed reward task, it is extremely hard for learning algorithms to find an optimal policy just according to these sparse training signals.
The additional training signals provided by SIL would significantly relieve this situation.
The key insight of SIL is that though the good past experiences of agents are not exactly expert, they can actually offer informative training signals for the agents.
Generally, SIL can be regarded as an extension to imitation learning.
The main difference from imitation learning is that imitation learning necessitates predefined expert demonstrations, which are prohibitively expensive and required much manual craft \cite{Multitask-imitation,DisagreementRegularizedIL}.
In SIL, the demonstrations are selected from its past experiences and the expert demonstration set will be filled gradually and update constantly.
Fig. \ref{intuition} presents our self-imitation learning framework for robots to learn how to behave.
The robot interacts with the environment and generates good or bad experiences. 
The experience data further guides the robot to perform better.
Consequently better experiences would be produced and it brings about a virtuous self-supervision circle between the robot's behaviors and historical experiences.
It is worth noting that in existing SIL methods, only good trajectories are collected to provide an auxiliary positive guidance for training parallelly with the standard RL training process.
However, our method includes bad trajectories as negative samples for training, 
since the positive samples would encourage the agents to approach them while the negative ones would prevent them from doing the same things.

In practice, SIL method is usually implemented on top of a RL framework, such as advantage actor-critic (A2C) \cite{SIL} or proximal policy optimization (PPO) \cite{GASIL}.
Recently, maximum entropy reinforcement learning attracts more attention for its impressive performance \cite{sac}.
Therefore, soft actor-critic (SAC), a state-of-the-art RL algorithm under maximum entropy reinforcement learning framework is used as the basis of our method.

\subsection{Maximum entropy reinforcement learning}
Reinforcement learning (RL) provides a powerful paradigm to learn optimal policy by trail-and-error for many complex control tasks. 
The RL agents observe the task environment in current time step $t$ to recover context state $s_{t} \in \mathcal{S}$ and decide an action $a_{t} \in \mathcal{A}$ to react to the corresponding state and have effect in its transition,
where $\mathcal{S}$ and $\mathcal{A}$ refer to states and actions set respectively.
The environment would response the agents an immediate reward $r_t$ indicating the credit of the action.
The goal of a standard model-free RL agent is to find a policy $\pi ^{*}\left (a_t|s_t \right )$ to maximize the expected sum of rewards over trajectories $\rho _{\pi}$:
\begin{equation}
\pi^{*} = \mathop{\arg\max}_{\pi} \sum_{t}^{\infty} \mathbb{E} \left [ r_{t}\left ( s_t, a_t \right ) \right ]
\end{equation}

The distribution of immediate rewards $r_t$ from environments is crucial for RL training efficiency.
Ideally, an immediate reward is expected to be obtained after the agent performs an action.
In this \emph{dense reward} situation, standard RL methods can learn quickly with these reasonable rewards.
However, most of the immediate rewards are absent in real-world robotic applications.
These sparse and delayed rewards make the learning efficiency of standard RL methods reducing significantly.
The extreme case is \emph{episodic rewards} where all the immediate rewards are zero and the only one informative signal is produced when an episode terminates, evaluating the performance of the agent in this trajectory.

The maximum entropy RL framework \cite{MaxEntropy} extends standard RL to find a soft policy with more stochasticity.
The extended objective differs from the standard with an additional regularization term of the policy entropy $\mathcal{H}$:
\begin{equation}
\label{MAXIMUM_ENTROPY}
\pi^{*} = \mathop{\arg\max}_{\pi} \sum_{t}^{\infty} \mathbb{E} \left [ r_{t}\left ( s_t, a_t \right )  + \alpha \mathcal{H}\left ( \pi \left ( \cdot|s_t\right ) \right ) \right ]
\end{equation}
where $\alpha$ is a temperature parameter controlling relative weight of entropy term in the whole objective, 
and the original objective of RL can be recovered in the limit as $\alpha \rightarrow 0$.
SAC provides a practical way to apply maximum entropy RL in continuous control tasks \cite{sac}.

\subsection{Training}
As mentioned above, not only the good experiences are important for learning, but the bad experiences are also informative.
The goodness of experiences is judged by their episodic rewards.
The trajectories with top-K highest episodic rewards will be preserved as good trajectories and all the other trajectories are bad.
More specifically, we maintain two experience replay buffers $\mathcal{D}_O$ and $\mathcal{D}_E$.
$\mathcal{D}_O$ is called \emph{\textbf{O}nline replay buffer}, in which we store all the experience trajectories $\tau = \left\{ s_t, a_t, r_t, s_{t+1} \right\}_{t=1}^{T}$ including good and bad.
$\mathcal{D}_E$ is called \emph{\textbf{E}xpert replay buffer}, in which we store good trajectories.
Considering the balance between positive and negative samples, the proportion of sample data from $\mathcal{D}_O$ and $\mathcal{D}_E$ is set to $1:1$ while replaying them.
This is crucial to make the training more efficient and stable \cite{SQIL}.

All the immediate rewards of the transition data $\phi_i = \left\{ s_i, a_i, r_i, s_{i+1} \right\}$ in these two experience replay buffers are fixed with constant values as follows:

\begin{equation}
	r_i=\left\{\begin{matrix}
 0,& \phi_i \in \mathcal{D}_O  \\ 
 1,& \phi_i \in \mathcal{D}_E
\end{matrix}\right.
\end{equation}
This mechanism of constant rewards is introduced in imitation learning to discriminate expert data and interaction data \cite{SQIL}.
In our self-imitation setting we use it to replace the assignment of the environmental temporal dynamics.
In this way, even though  the immediate rewards $r_i$ obtained from environments are sparse and delayed, transitions in the trajectories used for training could be guided properly.

According to SAC framework, the experience data will be replayed for the training of soft Q function and maximum entropy policy.
We here consider parametrized soft Q function $Q_{\theta}$ and policy $\pi_{\varphi}$ with parameter matrices $\theta$ and $\varphi$ correspondingly.
$Q_{\theta}$ is trained to minimize the soft Bellman residual:
\begin{equation}
\label{E1}
\begin{split}
J_{Q}\left (\theta \right ) = \mathbb{E}_{\tau \sim \mathcal{D}_{O}}\left [\frac{1}{2} \left ( Q_{\theta}\left (s_t, a_t \right ) - y \right ) ^2 \right ] 
\\
+ \mathbb{E}_{\tau \sim \mathcal{D}_E}\left [\frac{1}{2} \left (Q_{\theta}\left (s_t, a_t \right ) - y \right ) ^2 \right ]
\end{split}
\end{equation}
with
\begin{equation}
\begin{split}
y = r + \gamma \left ( \hat{Q}_{\bar{\theta}}\left( s_{t+1}, a_{t+1} \right) - \alpha \log \left(a_{t+1}|s_{t+1}\right ) \right),
\\
a_{t+1} \sim \pi_{\varphi} (\cdot | s_{t+1})
\end{split}
\end{equation}
where $\hat{Q}$ is the target soft Q function with delayed parameters $\bar{\theta}$ and 
$\alpha$ is entropy temperature parameter (see Equation \ref{MAXIMUM_ENTROPY}).
$r \in \left \{0, 1\right \}$ is the constant reward described above.
$\pi_{\varphi}$ is trained to maximize the estimation of soft Q value with additional entropy objective:
\begin{equation}
\label{E2}
\begin{split}
J_{\pi} \left( \varphi \right) = \mathbb{E}_{s_t \sim \mathcal{D}_{O}} \left[\alpha  \log \pi_{\varphi} \left( a_t|s_t\right) - Q_{\theta}\left( s_t, a_t \right)\right]
\\
+ \mathbb{E}_{s_t \sim \mathcal{D}_{E}} \left[ \alpha \log \pi_{\varphi} \left( a_t|s_t\right) - Q_{\theta}\left( s_t, a_t \right)\right],\\
a_{t} \sim \pi_{\varphi} (\cdot | s_{t})
\end{split}
\end{equation}

The training procedure is summarized in Algorithm \ref{pseudo}.
$\mathcal{D}_E$ is implemented by a priority queue, in which the subsequent better experience data will overwrite the priors, 
keeping $\mathcal{D}_E$ as the best performance the agent can reach.
$\mathcal{D}_O$ is a sequential queue with a standard replay buffer structure.
The size of $\mathcal{D}_E$ is determined by a hyperparameter  $\mathcal{M}$.
$\mathcal{M}$ controls how many good experiences the agent will imitate.
The full ablation study about $\mathcal{M}$ will be discussed in Section \ref{ablation}.
$\mathcal{D}_E$ and $\mathcal{D}_O$ are both empty at the beginning of training.
In the main loop we only use episodic rewards to inspect the quality of trajectories.
That is, no immediate rewards are involved in training.
The training batches $B_O$ and $B_E$ are sampled from $\mathcal{D}_O$ and $\mathcal{D}_E$ correspondingly with the same size.
Then they are concatenated into a single batch feeding to the networks. 
The implementation of SILCR can be found in \url{https://github.com/gouxiangchen/SILCR}.

\begin{algorithm}
	\caption{SILCR}
	\begin{algorithmic}[1]
		\Require Online replay buffer $\mathcal{D}_O \leftarrow \varnothing$, expert replay buffer $\mathcal{D}_E \leftarrow \varnothing$, Q function $Q_{\theta_0}$, policy $\pi_{\varphi_0}$
		\For {$i = 0,1,2...$}
			\State perform $\pi_{\varphi_i}$ to generate $good$ and $bad$ trajectories
			\State Store $\tau_{bad}$ and $\tau_{good}$ in $\mathcal{D}_O$ with $r_t=0$
			\State Store $\tau_{good}$ in $\mathcal{D}_E$ with $r_t=1$
			\State Randomly sample data $B_O$ from $\mathcal{D}_O$ 
			\State Randomly sample data $B_E$ from $\mathcal{D}_E$, size in $|B_O|$
			\State Training batch $B \leftarrow B_O \cup B_E $
			\State $\theta_{i+1} \leftarrow \theta_{i} - \eta_{Q} \nabla J_{Q}(\theta)$ (See Equation \ref{E1})
			\State $\varphi_{i+1} \leftarrow \varphi_{i} - \eta_{\pi} \nabla J_{\pi}(\varphi)$ (See Equation \ref{E2})
		\EndFor
	\end{algorithmic}
	\label{pseudo}
\end{algorithm}

\section{Evaluation}
We empirically evaluate our method to answer the following questions: (1) Can SILCR proceed successfully and find feasible control policy in challenging robotic control tasks with sparse rewards? (2) How well is the performance of SILCR compared with other baseline methods with dense rewards or expert demonstrations? (3) How the hyperparameter $\mathcal M$, the size of expert replay buffer, influences performance of our method?

\begin{figure*}[thpb]
\centering
\includegraphics[width=0.96\textwidth]{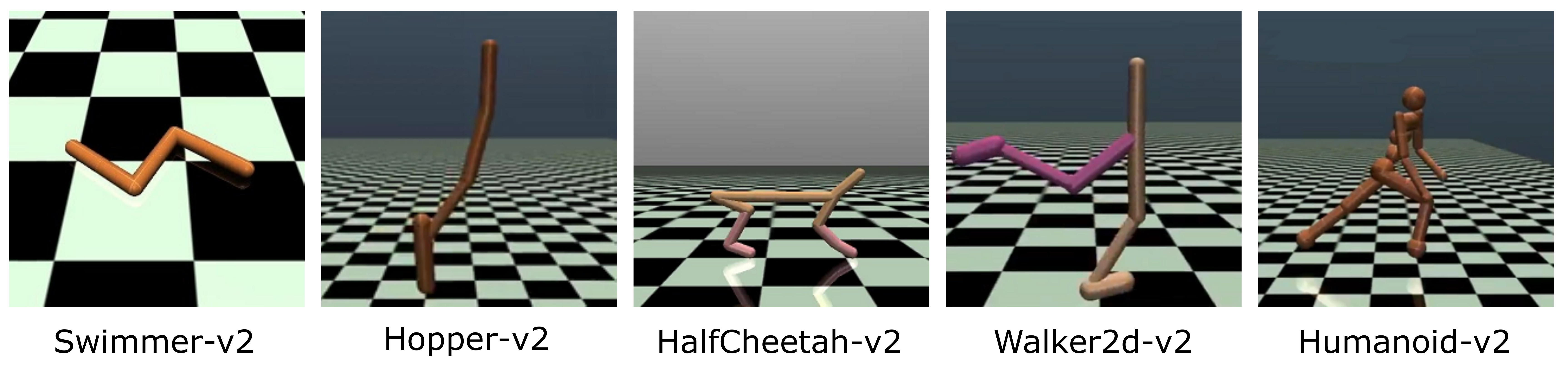} 
\caption{Five continuous robotics control tasks in MuJoCo simulation.}
\label{mujoco}
\end{figure*}

As illustrated in Fig. \ref{mujoco}, five continuous robotics control tasks of OpenAI Gym MuJoCo \cite{openAIGym,MuJoCo} are chosen as our benchmarks.
In these tasks, robots are asked to learn how to autonomously swim forward, jump, stand up, walk and etc.
The state dimensions of these tasks are from 8 to 376 and the action dimensions are from 2 to 17.
Among them, the \emph{Humanoid-v2}, in which a humanoid robot is required to stand steadily and walk forward, has been demonstrated as a very complex and high-dimension challenging task for robotic learning \cite{RLBenchmark}.

We consider following four methods for comparison:
\begin{itemize}
\item \textbf{SAC}: a baseline soft actor-critic algorithm \cite{sac}, the state-of-the-art model-free reinforcement learning method.
\item \textbf{GASIL}: a generative adversarial extension of self-imitation learning.
\item \textbf{SQIL}: soft Q imitation learning algorithm \cite{SQIL}, a recent imitation learning showing excellent performance.
\item \textbf{SILCR}: our self-imitation learning method with only episodic reward available.
\end{itemize}
All the hyperparameters used in our implementation are presented in TABLE \ref{hyperparameter}.
It should be noted that the values of these hyperparameters are common in original SAC algorithm except \emph{Expert buffer size}.
The scales of this hyperparameter will be discussed in Section \ref{ablation}.
To reduce the influence of random error, all the experiments are averaged over ten runs with different random seeds.

\begin{table}[h]
\centering
\caption{Hyperparameters used cross all tasks. }
\label{hyperparameter}
\vspace{1em}
\begin{tabular}{@{}lc@{}}
\toprule	
\textbf{Hyperparameter}     & \textbf{Value}                                                                   \\ \midrule
Actor learn rate            & $3\times 10^{-4}$                                                                \\
Critic learn rate           & $3\times 10^{-4}$                                                                \\
Optimizer                   & Adam                                                                             \\
Expert buffer size          & $5\times 10^{4}$                                                                 \\
Online buffer size          & $10^{6}$                                                                 \\
Discount factor ($\gamma$)  & 0.99                                                                             \\
Target update rate ($\tau$) & 0.05                                                                             \\
Batch size                  & 128                                                                              \\
Actor network architecture  & \begin{tabular}[c]{@{}c@{}}(state dim, 800, \\ 400, action dim)\end{tabular}     \\
Critic network architecture & \begin{tabular}[c]{@{}c@{}}(state dim + action dim, \\ 800, 400, 1)\end{tabular} \\ 
\bottomrule
\end{tabular}

\end{table}

\begin{figure*}[thpb]
\centering
\includegraphics[width=0.95\textwidth]{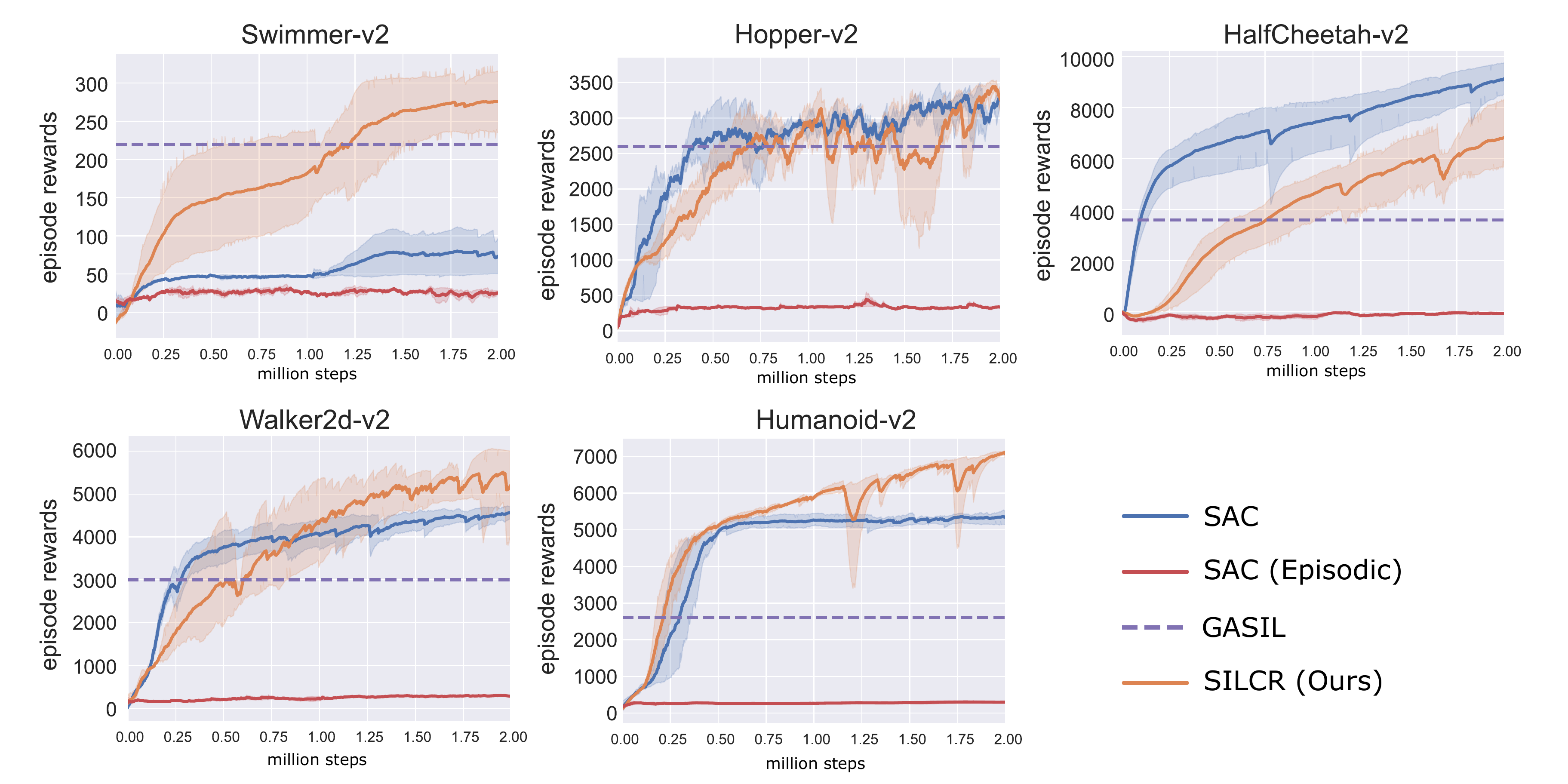} % Reduce the figure size so that it is slightly narrower than the column.
\caption{Learning curves on five MuJoCo continuous control benchmarks. The results of GASIL are intercepted from \cite{GASIL} at 2 million steps. The results are averaged over five runs with different random seeds.}
\label{Result}
\end{figure*}

\begin{figure*}[thpb]
\centering
\includegraphics[width=0.92\textwidth]{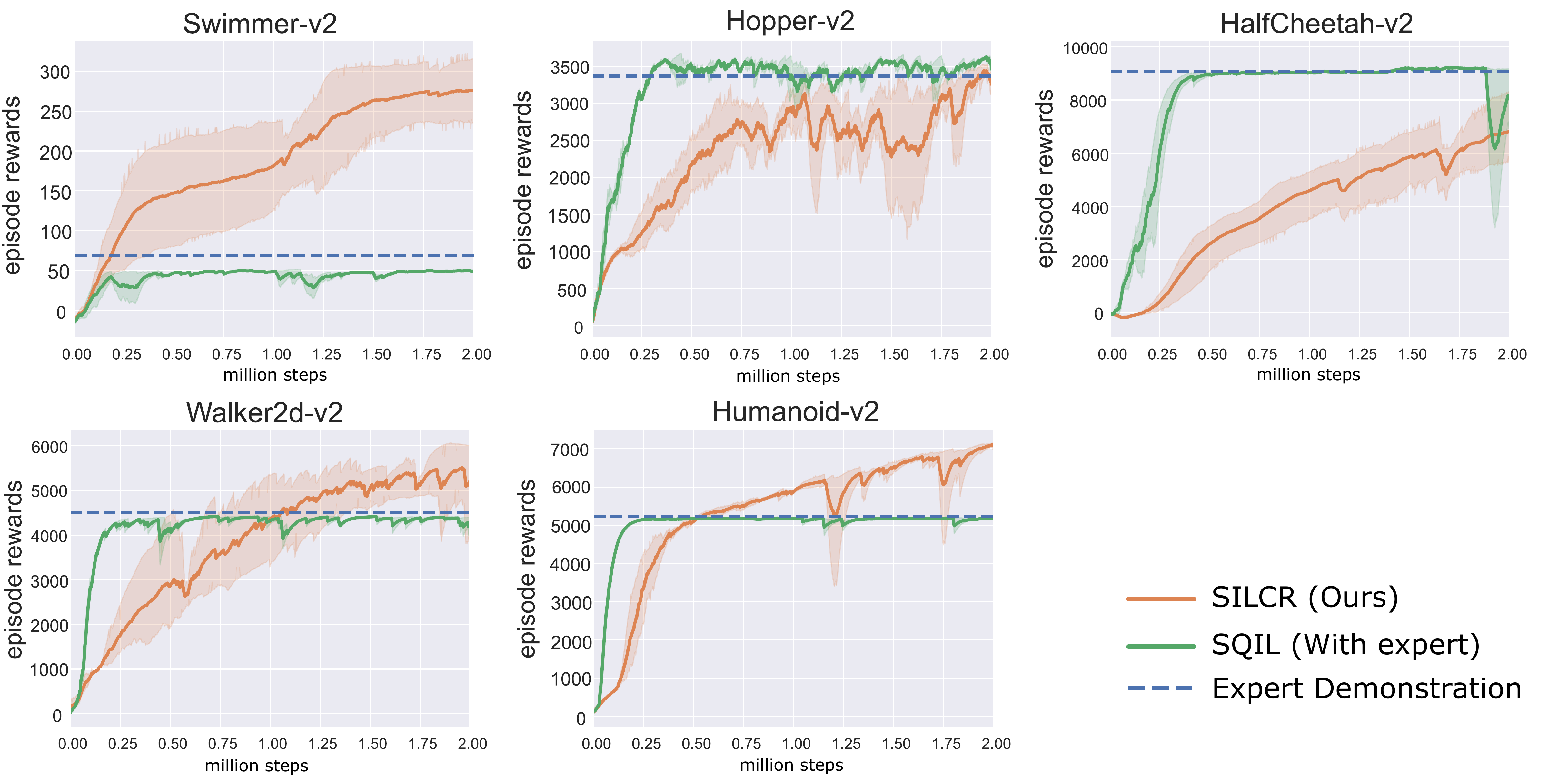} % Reduce the figure size so that it is slightly narrower than the column.
\caption{Comparison with imitation learning on five MuJoCo continuous control benchmarks. The expert is generated by a SAC agent trained in dense reward settings.}
\label{Result-il}
\end{figure*}

\subsection{Sparse reward settings}
In sparse reward settings, the immediate rewards obtained from environments are mostly zero.
Reference to common practice \cite{DiverseSIL}, we modify the original MuJoCo tasks into an episodic reward setting version.
That is, in episodic reward setting, all the immediate rewards at each timestep are set to zero along the entire episode and an episodic reward is provided at the last timestep of the episode.
We run SAC and SILCR in the modified tasks and the learning results are shown in Fig. \ref{Result}.
In this setting, SAC fails to learn anything within 2 million steps in all of the tasks and behaves randomly.
SILCR dramatically outperforms SAC.
In all the tested environments, SILCR exceeds SAC at the very beginning and learns in an impressive efficiency, reaching high episodic rewards quickly.
Since the official GASIL implementation is not publicly available, a fair reproduction of it in sparse reward settings is hardly possible.
Therefore, the comparison with GASIL is only conducted in dense reward settings as described below.

\subsection{Dense reward settings}
Standard RL methods have shown their ability to find optimal policy in difficult tasks when dense rewards are available.
We further conduct experiments to compare the performance and data efficiency of these methods in dense reward setting.
The learning results are shown in Fig. \ref{Result}.
In most tasks, SILCR exceeds baseline SAC and reaches higher episodic rewards in 2 million steps.
Specifically in \emph{Swimmer-v2}, SILCR surpasses SAC at the beginning of training.
SAC struggles to obtain a good performance while SILCR reaches an impressive final results.
In \emph{Humanoid-v2}, the behaviors of SAC and SILCR are similar in the early training, but SAC finally converges to a suboptimal result while SILCR can obtain higher levels.
It is worth noting that our method is still under episodic reward settings with high sparsity.

GASIL also requires dense rewards from environments.
For a fair comparison, we intercept the results of GASIL at 2 million steps from the original paper \cite{GASIL}.
The results show that GASIL can not follow the levels of baseline SAC in most tasks and SILCR outperforms it in all the test environments.

\subsection{Comparison with imitation learning}
\label{CompareIL}
We further compare our method with a representative imitation learning method SQIL.
Rather than learning from scratch, imitation learning needs to prepare expert data firstly.
In our experiment setting, the expert data for SQIL is generated by a SAC agent trained with dense rewards. 
The learning results are shown in Fig. \ref{Result-il}, where the dashed line indicates the level of expert behaviors.
Predictably, with the help of expert data, the training speed of imitation learning improves significantly.
However, the performance of imitation learning is limited largely by the quality of expert data and it can hardly exceed the performance of the expert.
In contrast, there has no access to the expert data in SILCR, and the imitation goals are improved constantly by the agent itself.
Consequently, SILCR is potential to find optimal policy as the number of training iterations goes up while imitation learning is constrained by the performance of expert.

\subsection{Ablation study over the size of expert replay buffer}
\label{ablation}
Deep neural network based methods introduce many  hyperparameters inevitably,
and the tuning of these non-trivial hyperparameters becomes a laborious task.
For example, training generative adversarial networks (GANs) is a notoriously challenging task
in which a significant amount of hyperparameters has to be tuned carefully in order to reach a good performance \cite{GANTricks}.
Moreover, small changes in hyperparameters may cause a dramatic degeneration in performance.
Therefore, a small amount and easy tuning of hyperparameters could make deep learning methods more practical and reproducible.

\begin{figure}[thpb]
\centering
\includegraphics[width=0.96\columnwidth]{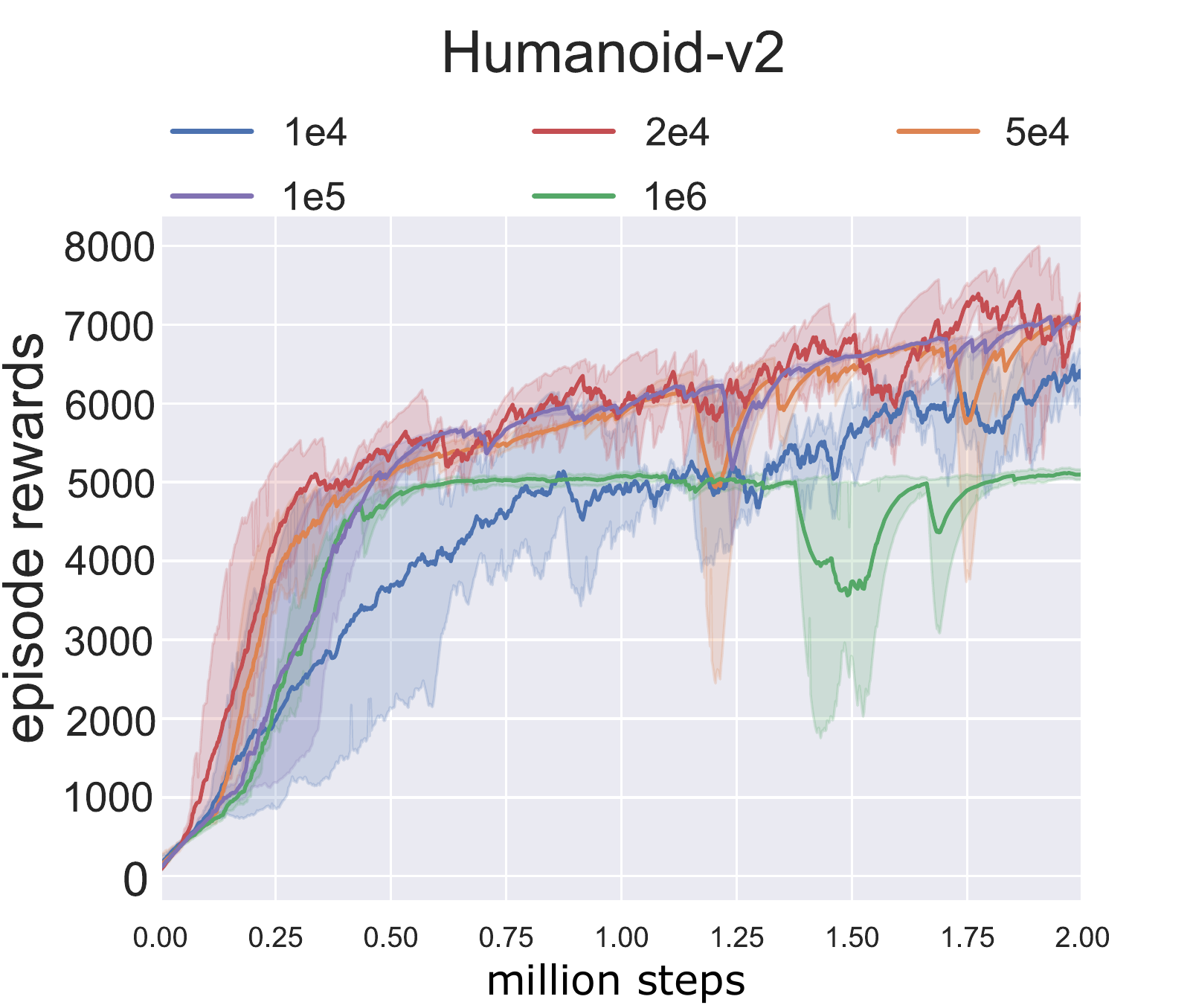}
\caption{Ablation study over the size of expert replay buffer $\mathcal{M}$. Five experiments are conducted to investigate the influence of the hyperparameter.}
\label{ablation_result}
\end{figure}

The hyperparameter introduced in our method is the size of expert replay buffer $\mathcal{M}$.
Intuitively, $\mathcal{M}$ should not be too large or too small.
A large buffer may reserve the poor experiences as expert for a long period while a small buffer may undermine the diversity of expert behaviors.
We conduct experiments on task \emph{Humanoid-v2} with five choices of expert buffer sizes of $1\times 10^4$, $2\times 10^4$, $5\times 10^4$, $1\times 10^5$,  $1\times 10^6$, where the smallest size of $1\times 10^4$ contains 10 of full trajectories.
The other hyperparameters are the same as in TABLE \ref{hyperparameter}.
The results are shown in Fig. \ref{ablation_result}.
The results show that the performances of SILCR with hyperparameters of $2\times 10^4$, $5\times 10^4$ and $1\times 10^5$ are very similar where all of them are potential to get episodic rewards more than $7000$.
The results indicate that a reasonable value from $2\times 10^4$ to $1\times 10^5$ could reach a good performance.
In the case of $1\times 10^4$, the agent tends to learn more slowly and with more variance at the beginning of training due to a small size of expert data, which leads to less diversity of the expert policy and reduces the learning efficiency.
In the another extreme case of $1\times 10^6$, the performance degrades to standard SAC level and reaches a suboptimal result with about $5000$ episodic rewards.
This is because that many poor experiences are still in expert buffer after many steps of training, which misleads the agent.

\section{Conclusions and Future Works}
In this paper, we presented a self-imitation learning method to enable robots to learn behavior skills in environments with sparse and delayed rewards.
We address the limitations in existing self-imitation learning methods by using a constant reward assignment to relieve the credit assignment problem in sparse reward settings.
The experiments on robotic continuous control tasks demonstrate the effectiveness of our method.
Furthermore, our method can be regarded as an extension to imitation learning in the sense that it treats the agents' own good past experiences as the imitated expert.
However, it avoids the main drawback of imitation learning since there is no need for manually defined expert demonstrations.
The results show that our method is potential to find optimal policy and surpass expert levels.
Though we demonstrate our method in continuous robotic control tasks, it is suitable for other environments in sparse reward settings.

In the future, we would like to extend our method to different model-free RL algorithms such as DQN algorithm family \cite{dqn} in discrete action space terms.
It is an interesting direction to investigate the warm start of $\mathcal{D}_{E}$, combining with state-of-the-art exploration methods, such as RND \cite{Burda2019ExplorationBR}, to obtain several high episodic reward trajectories in the beginning of training.

\bibliographystyle{IEEEtran}
\bibliography{reference}

\end{document}